\documentclass[runningheads]{llncs}
\usepackage{graphicx}
\usepackage{subfig,multirow}
\usepackage[ruled]{algorithm2e}
\usepackage{amsmath,amssymb}

\usepackage[T1]{fontenc}
\usepackage{graphicx}
\begin{document}
\title{Cluster Contrast for Unsupervised Person Re-Identification}
%
%
\author{Zuozhuo Dai\inst{1} \and
Guangyuan Wang\inst{1} \and
Weihao Yuan\inst{1} \and Siyu Zhu\inst{1} \and Ping Tan\inst{2}}
\authorrunning{Z. Dai et al.}
%
\institute{Alibaba Cloud \and Simon Fraser University}
\footnotetext[1]{Siyu Zhu is the corresponding author} 
\maketitle              
\begin{abstract}
Thanks to the recent research development in contrastive learning, the gap of visual representation learning between supervised and unsupervised approaches has been gradually closed in the tasks of computer vision.
In this paper, we focus on the downstream task of unsupervised person re-identification (re-ID).
State-of-the-art unsupervised re-ID methods train the neural networks using a dictionary-based non-parametric softmax loss. They store the pre-computed instance feature vectors inside the dictionary, assign pseudo labels to them using clustering algorithm, and compare the query instances to the cluster using a form of contrastive loss.
To enforce a consistent dictionary, that is the features in the dictionary are computed by a similar or the same encoder network,  we present Cluster Contrast which stores feature vectors and computes contrastive loss at the cluster level.
Moreover, the momentum update is introduced to reinforce the cluster-level feature consistency in the sequential space.
Despite the straightforward design, experiments on four representative re-ID benchmarks demonstrate the effective performance of our method.
\end{abstract}
\section{Introduction}
\begin{figure}[!t]
    \centering
    \includegraphics[width=0.45\linewidth]{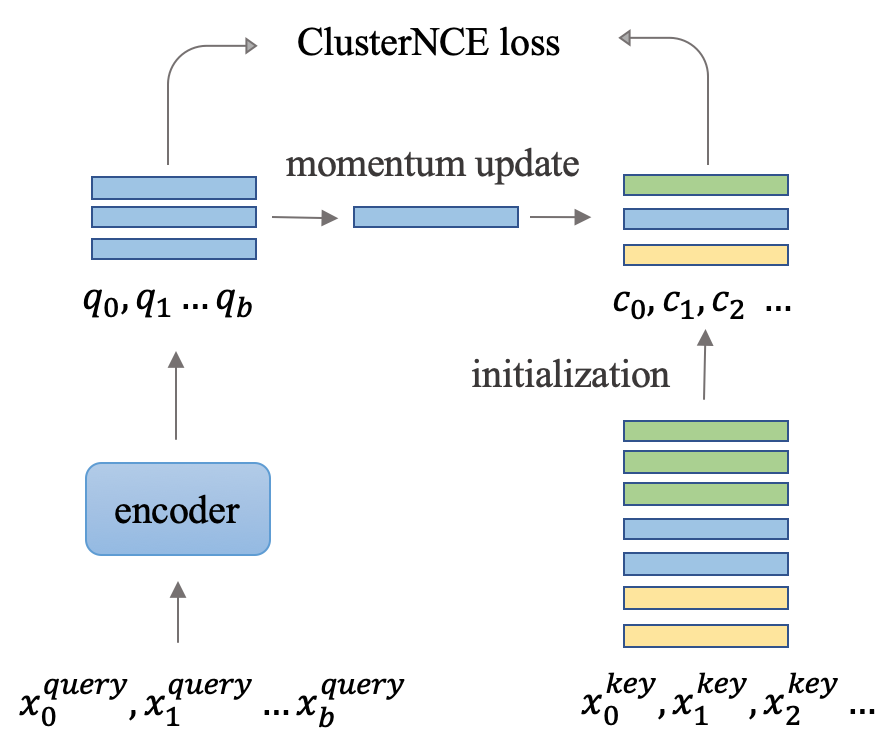}
    \caption{The Cluster Contrast computes the contrastive loss in cluster level with momentum update. In the cluster level memory dictionary, the cluster feature is initialized as the averaged feature of the corresponding cluster and updated by the batch query instance. $x \in X$ is the training dataset. $q$ is the query instance feature vector. $c_k$ stands for the $k$-th cluster feature vector. 
    Feature vectors with the same color belong to the same cluster.}  
    \label{fig:clustercontrastloss}
\end{figure}

Deep unsupervised person re-ID aims to train a neural network capable of retrieving a person of interest across cameras without any labeled data.
This task has attracted increasing attention recently due to the growing demands in practical video surveillance and the expensive labeling cost.
There are mainly two approaches to address this problem.
One is the purely unsupervised learning person re-ID, which generally exploits pseudo labels from the unlabeled data~\cite{fan2018unsupervised,fu2019self,ge2020self,lin2019bottom,wang2020unsupervised}.
The other is the unsupervised domain adaptation person re-ID, which first pre-trains a model on the source labeled dataset, and then fine-tunes the model on the target unlabeled dataset~\cite{deng2018image,lin2018multi,wang2018transferable,wei2018person,yu2019unsupervised,zhong2018generalizing,zhong2019invariance}.
Generally, the performance of domain adaptation is superior to that of unsupervised learning because of the introduction of the external source domain.
However, domain adaptation still suffers from the complex training procedure and requires that the difference between the source and target domain is not significant. In this paper, we focus on learning the person re-ID task without any labeled data, namely the purely unsupervised learning. 

Recently, the unsupervised representation learning methods~\cite{bachman2019learning,caron2020unsupervised,chen2020simple,chen2020improved,hadsell2006dimensionality,he2020momentum,henaff2020data,hjelm2018learning,misra2020self,van2018representation,tian2020contrastive,Wu_2018_CVPR,ye2019unsupervised} with contrastive loss~\cite{hadsell2006dimensionality} have gradually closed the performance gap with supervised pretraining in computer vision. 
Here, the contrastive loss~\cite{hadsell2006dimensionality} aims to compare pairs of image features so that the positive sample pairs are pulled together and the negative sample pairs are pulled away. 
Specifically, InstDisc~\cite{Wu_2018_CVPR} proposes an instance level memory bank for instance discrimination. 
It compares the query image features to all the instance features in the memory bank. 
Subsequently, the MoCo series~\cite{chen2020improved,chen2021empirical,he2020momentum} highlight the consistent memory dictionary in contrastive learning of visual representations.
MoCo approximates the contrastive loss by sampling a subset of instances in the memory dictionary and uses the momentum-based moving average of the query encoder. 
Meanwhile, SimCLR~\cite{chen2020simple} uses a large enough batch size to compute contrastive loss, which requires hundreds of TPU cores.
Later on, SwAV~\cite{caron2020unsupervised} computes the contrastive loss in cluster level.
It enforces the cluster assignment results rather than comparing sampling instance features. 
Since the cluster number is fixed in online clustering, SwAV does not require the large instance feature memory bank or large batch size to enforce the feature consistency. 

Inspired by the great success of contrastive learning, recent works~\cite{chen2021ice,ge2020self,wang2020unsupervised,wang2021camera,fan2018unsupervised,fu2019self,lin2019bottom,xuan2021intra,zheng2021online,zhang2021refining} try to apply such ideology to the downstream re-ID tasks. 
In more details, such approaches exploit the memory dictionary and pseudo labels from clustering to train the neural network. 
At the beginning of each epoch, all the image features of the training data are extracted by the current neural network. 
Then, such image features are stored in a memory dictionary and a clustering algorithm, like DBSCAN~\cite{ester1996density} or K-means~\cite{macqueen1967some} is employed to cluster image features and produce pseudo labels.
Meanwhile, the cluster ID is assigned to each image as the person identity.
Finally, the neural network is trained with a contrastive loss such as triplet loss~\cite{hermans2017defense,schroff2015facenet}, InfoNCE loss~\cite{oord2018representation}, or other non-parametric classification loss~\cite{wang2020unsupervised} between the feature vectors of every instance inside the memory dictionary and the query instance.
Since the instance features updated in one iteration are limited by the batch size, the instance features from the newly updated encoder network are not consistent with the previous ones.
This problem of feature inconsistency in memory dictionary is especially serious in large-scale re-ID datasets like MSMT17~\cite{wei2018person}.

To enforce a consistent feature dictionary, we propose Cluster Contrast for unsupervised person re-ID.
Remarkably, the ideology of Cluster Contrast is inspired by the contrasting cluster assignment technique from SwAV~\cite{caron2020unsupervised}.
Different from SwAV which adopts an online clustering approach with a fixed number of clusters, we use an offline clustering method~\cite{fan2018unsupervised,ge2020self} which demonstrates superior clustering performance in re-ID tasks  and remove un-clustered outliers.
Then, a cluster-level memory dictionary is built and each dictionary key corresponds to a cluster which is represented by a single feature vector.
More specifically, this cluster feature is initialized as the average feature of all the images from the same cluster and updated by the batch query instance features during training.  
Accordingly, we propose a cluster-level InfoNCE loss, denoted as ClusterNCE loss, which computes contrastive loss between cluster feature and query instance feature as illustrated in Figure \ref{fig:clustercontrastloss}.  
Moreover, we apply the ideology of momentum update policy from MoCo~\cite{he2020momentum} to the cluster level memory to further boost the feature consistency of cluster representations in the sequential space.



In summary, our proposed Cluster Contrast for unsupervised re-ID  has the following contributions:
\begin{itemize}
	\item We introduce the cluster-level memory dictionary which initializes, updates, and performs contrastive loss computation at the cluster level. The cluster feature embedding helps to alleviate the feature inconsistency problem.
	\item We apply the momentum updating policy to the cluster feature representation and further enforce the feature consistency in the memory dictionary. 
 	\item We demonstrate that the proposed unsupervised approach with Cluster Contrast achieves state-of-the-art performance on three purely unsupervised re-ID benchmarks. 
\end{itemize}

\section{Related Work}

\paragraph{Deep Unsupervised Person Re-ID.}
Deep unsupervised person re-ID can be summarized into two categories.
The first category is unsupervised domain adaptation re-ID, which utilizes transfer learning to improve unsupervised person re-ID~\cite{ge2020self,wang2020unsupervised,deng2018image,ge2020mutual,isobe2021towards,zhai2020ad,lin2018multi,wang2018transferable,wei2018person,yu2019unsupervised,zhong2018generalizing,zhong2019invariance}. The second category is pure unsupervised learning person re-ID~\cite{fan2018unsupervised,fu2019self,lin2019bottom,chen2021ice,wang2021camera,xuan2021intra,zheng2021online,zhang2021refining}, which trains model directory on unlabeled dataset. State-of-the-art unsupervised learning re-ID pipeline generally involves three stages: memory dictionary initialization, pseudo label generation, and neural network training. 
Previous works have made significant improvements either in parts or on the whole pipeline.
Specifically, Lin\textit{et al.}~\cite{lin2019bottom} treats each individual sample as a cluster, and then gradually groups similar samples into one cluster to generate pseudo labels. MMCL~\cite{wang2020unsupervised} predicts quality pseudo labels comprising of similarity computation and cycle consistency. It then trains the model as a multi-classification problem. SPCL~\cite{ge2020self} proposes a novel self-paced contrastive learning framework that gradually creates more reliable cluster to refine the memory dictionary features. OPLG~\cite{zheng2021online} and RLCC~\cite{zhang2021refining} explore the temporal label consistency for better pseudo label quality. In addition to pseudo label, another stream of camera-aware methods~\cite{wang2021camera,chen2021ice} utilizes camera ID information as additional supervision signal to further improve the unsupervised re-ID performance. In this paper, we focus on purely unsupervised person re-ID, but our method can be easily generalized to unsupervised domain adaptation and camera-aware methods.

\paragraph{Memory Dictionary.}
Contrastive learning~\cite{bachman2019learning,caron2020unsupervised,chen2020simple,chen2020improved,hadsell2006dimensionality,he2020momentum,henaff2020data,hjelm2018learning,misra2020self,van2018representation,tian2020contrastive,Wu_2018_CVPR,ye2019unsupervised} can be thought of as training an encoder for a dictionary look-up task. Since it is too expensive in both memory and computation to compare all the image pairs within a dataset, several recent studies~\cite{bachman2019learning,he2020momentum,henaff2020data,hjelm2018learning,oord2018representation,tian2019contrastive,wu2018unsupervised,zhuang2019local} on unsupervised visual representation learning present promising results through building dynamic dictionaries. Moco~\cite{he2020momentum} builds a memory dictionary as a queue of sampled images. The samples in memory dictionary is replaced consistently on the fly to keep the feature consistency with the newly updated model. SimCLR~\cite{chen2020simple} shows that the instance memory can be replaced by a large enough batch of instances. Similar to unsupervised visual representation learning, state-of-the-art unsupervised person re-ID methods also build memory dictionaries for contrastive learning~\cite{Xiao_2017_CVPR,wang2020unsupervised,ge2020mutual,ge2020self}. During training, instance feature vectors in the memory dictionary are updated by the corresponding query instances features. Recently, SwAV~\cite{caron2020unsupervised} proposes an efficient online clustering method which approximates the contrastive loss of all image pairs by clustering centers, without requiring a large batch size or large memory bank. Inspired by SwAV~\cite{caron2020unsupervised} and Moco~\cite{he2020momentum}, we apply the ideology of cluster level contrastive learning and momentum update to the downstream unsupervised re-ID tasks and alleviate the problems of the large memory bank and memory dictionary inconsistency. Unlike SwAV in which the number of clusters is fixed, the proposed Cluster Contrast gradually selects reliable labels and dynamically refines the clustering results during training.  

\paragraph{Loss Functions.}
In supervised person re-ID, the batch hard triplet loss has proved to be effective solutions to improve the re-ID performance~\cite{chen2019self,dai2019batch,dai2018batch,guo2019beyond,liu2018pose,song2019generalizable,wang2018mancs,zhang2019densely,zhou2019discriminative}. In unsupervised person re-ID, since there is no ground truth person identity and the pseudo labels are changing during training, non-parametric classification loss such as InfoNCE~\cite{oord2018representation} are used as identity loss. Similar to InfoNCE, Tong\textit{et al.}~\cite{Xiao_2017_CVPR} designs an Online Instance Matching (OIM) loss with a memory dictionary scheme which compares query image to a memorized feature set of unlabelled identities. Wang and Zhang~\cite{wang2020unsupervised} introduce the memory-based non-parametric multi-label classification loss (MMCL), which treat unsupervised re-ID as a multi-label classification problem. In order to mitigate noisy pseudo labels, MMT~\cite{ge2020mutual} proposes a novel soft softmax-triplet loss to support learning with soft pseudo triplet labels. SPCL~\cite{ge2020self} introduces a unified contrastive loss including both source domain dataset and target domain dataset. In this paper, we apply InfoNCE loss between cluster feature and query instance feature on unsupervised re-ID.
\begin{figure}[!t]
    \centering
    \includegraphics[width=1.0\linewidth]{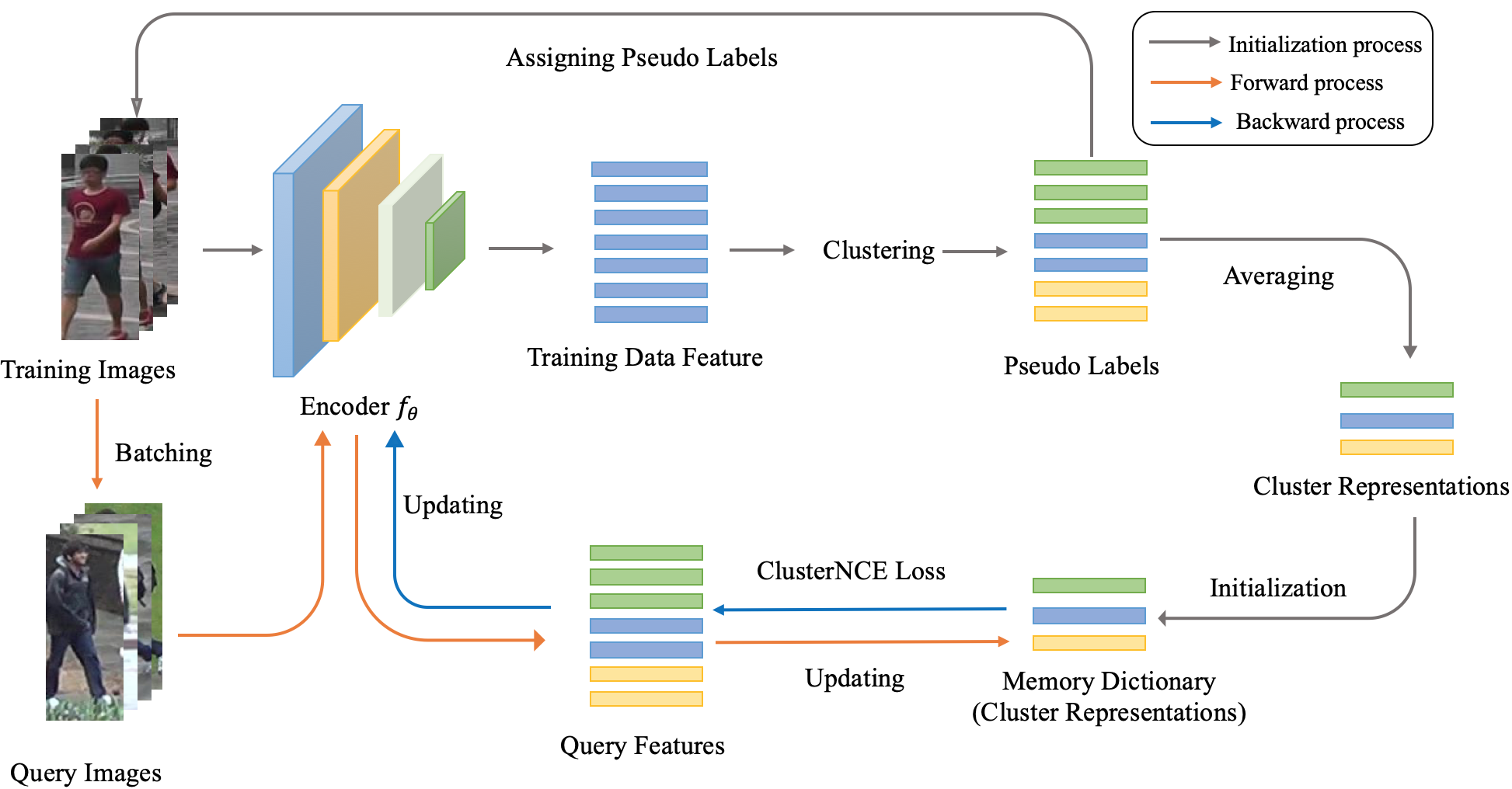}
    \caption{The system pipeline of our unsupervised person re-ID method. The upper branch depicts the memory initialization stage. The training data features are assigned pseudo labels by clustering, where features of the same color belong to the same cluster. The lower branch represents the model training stage. Query features in iterative mini-batch are used to update the memory cluster representations with a momentum. The ClusterNCE loss computes the contrastive loss between query features and all cluster representations.}
    \label{fig:pipeline}
\end{figure}

\section{Method}
We first introduce our overall approach at a high level in Section~\ref{sec:overview}.
Then, we compare the multiple contrastive learning approaches for person re-ID with our proposed cluster contrast method in Section~\ref{sec:ContrastiveLearning}.
Finally in Section~\ref{sec:ClusterContrast}, we explain the details of momentum update in Cluster Contrast along with its working theory.

\subsection{Overview}
\label{sec:overview}

State-of-the-art unsupervised learning methods~\cite{wang2020unsupervised,ge2020mutual,ge2020self,chen2021ice} solve the unsupervised learning person re-ID problem with contrastive learning. 
Specifically, they build a memory dictionary that contains the features of all training images. 
Each feature is assigned a pseudo ID generated by a clustering algorithm. 
During training, the contrastive loss is minimized to train the network and learn a proper feature embedding that is consistent with the pseudo ID.

We focused on designing a proper contrastive learning method to keep the whole pipeline simple while obtaining better performance.
An overview of our training pipeline is shown in Figure~\ref{fig:pipeline}.
The memory dictionary initialization is illustrated in the upper branch.
We use a standard ResNet50~\cite{he2016deep} as the backbone encoder which is pretrained on ImageNet to extract feature vectors, and has basic discriminability though not optimized for re-ID tasks.
We then apply the DBSCAN~\cite{ester1996density} clustering algorithms to cluster similar features together and assign pseudo labels to them. 
The cluster feature representation is calculated as the mean feature vectors of each cluster. 
The memory dictionary is initialized by these cluster feature representations and their corresponding pseudo labels. 
As shown in the lower branch, during the training stage, we compute the ClusterNCE loss between the query image features and all cluster representations in the dictionary to train the network. 
Meanwhile, the dictionary features are updated with a momentum by the query features.

To facilitate the description of methods, we first introduce the notations used in this paper.
Let $X = \{x_1,x_2, \dots, x_N\}$ denote the training set with $N$ instances. 
And $U=\{u_1,u_2,\dots,u_n \}$ denotes the corresponding features obtained from the backbone encoder $f_\theta$, described as $u_i = f_\theta (x_i)$.
$q$ is a query instance feature extracted by $f_\theta (\cdot)$, where the query instance belongs to $X$.

\begin{figure}[!t]
    \subfloat[Multi-label classification loss]{
        \centering
		\includegraphics[height=27mm]{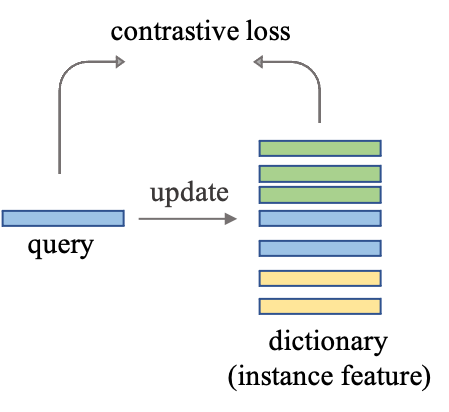}
    }
    \hfill
    \subfloat[Instance level InfoNCE loss]{
        \centering
		\includegraphics[height=27mm]{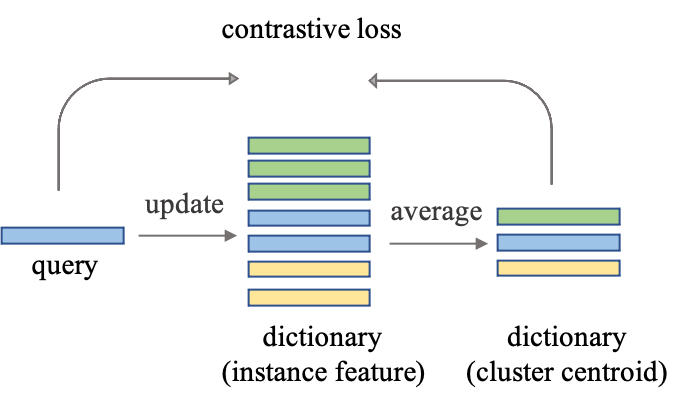}
    }
    \hfill
    \subfloat[ClusterNCE loss (ours)]{
        \centering
		\includegraphics[height=29mm]{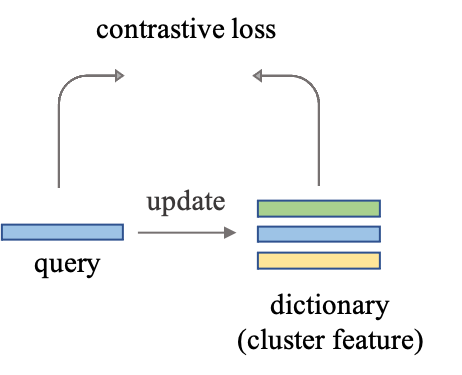}
    }
	\caption{Comparison of three types of memory-based non-parametric contrastive learning losses for re-ID. Different color features indicate different clusters. (a) computes the loss and updates the memory dictionary both at the instance level~\cite{wang2020unsupervised}. (b) computes the loss at the cluster level but updates the memory dictionary at the instance level~\cite{ge2020self}. (c) is our proposed approach and it computes the loss and updates the memory dictionary both at the cluster level.}
	\label{fig:loss}
\end{figure}

\subsection{Cluster Contrast}
\label{sec:ContrastiveLearning}

In this section, we analyze different contrastive learning methods to motivate our design of Cluster Contrast. As shown in Figure~\ref{fig:loss} (a), the multi-label classification loss computes the loss in the instance level through an instance-wise contrastive loss. It stores all image feature vectors in the memory dictionary and computes multi-class score by comparing each query feature to all of them. The memory dictionary is updated by the query features after each training iteration.


In Figure~\ref{fig:loss} (b), SPCL~\cite{ge2020self} computes the loss at cluster level through a cluster-wise InfoNCE loss. It can be defined as follows:

\begin{equation}\label{eq:SPCLLoss}
    L_q = -\log\frac{exp(q \cdot c_+/\tau)}{\sum_{k=1}^K{exp(q \cdot c_k/\tau)}}
\end{equation}
where $\tau$ is a temperature hyper-parameter,  $\{c_1, c_2, \dots, c_K\}$ are the cluster centroids and $K$ stands for the number of clusters. 
It uses the cluster centroid as the cluster level feature vector to compute the the distances between query instance $q$ and all the clusters. $c_+$ is the positive cluster feature which $q$ belongs to.
The cluster centroids are calculated by the mean feature vectors of each cluster as:
\begin{equation}\label{eq:SPCLUpdate_c}
    c_k = \frac{1}{|\mathcal{H}_k|}\sum_{u_i \in \mathcal{H}_k}u_i
\end{equation}
where $\mathcal{H}_k$ denotes the $k$-th cluster set and $|\cdot|$ indicates the number of instances per cluster. 
$\mathcal{H}_k$ contains all the feature vectors in the cluster $k$.
But similar to multi-classification loss , it stores all image feature vectors in the memory dictionary. The stored image feature vectors are then updated by corresponding query image feature.

\begin{figure}[t]
    \centering
    \includegraphics[width=0.8\linewidth]{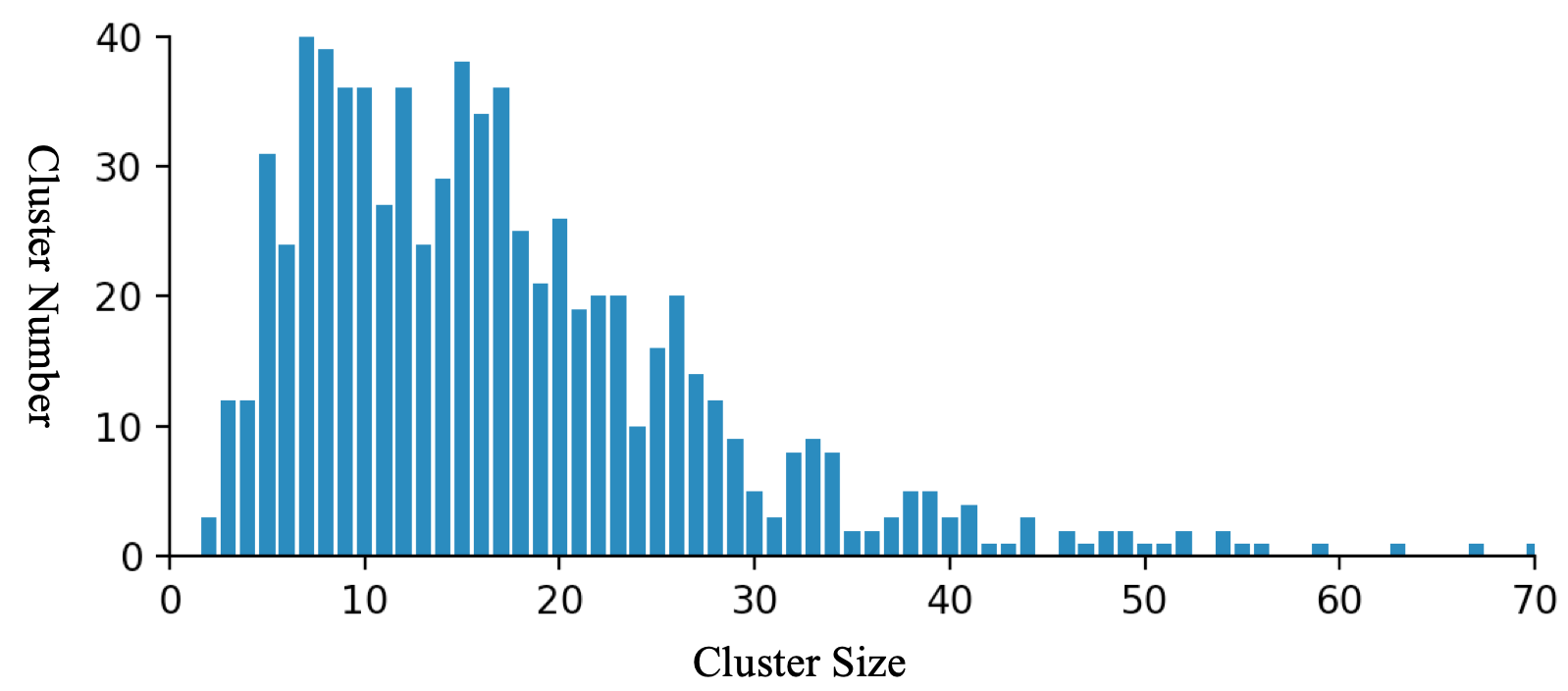}
    \caption{The cluster size follows a normal distribution in Market1501 dataset.}
    \label{fig:clustersize}
\end{figure}

Both Figure~\ref{fig:loss} (a) and Figure~\ref{fig:loss} (b) update the feature vectors at an instance level, resulting in feature inconsistency problem.
As shown in Figure~\ref{fig:clustersize}, the cluster size is unbalancedly distributed.
In every training iteration, in a large cluster only a small fraction of the instance features can be updated due to the batch size limitation, whereas in a small cluster all the instances can be updated.
Thus, the updating process is highly varied, and the contrastive loss computed by comparing all instance features is not consistent with the newest model. In each iteration, the network is constantly updated, which causes inconsistent oscillatory distribution of mini-batches.
In contrast, we design our ClusterNCE loss as shown in Figure~\ref{fig:loss} (c) using the following equation:
\begin{equation}\label{eq:ClusterNCELoss}
    L_q = -\log\frac{exp(q \cdot \phi_+/\tau)}{\sum_{k=1}^K{exp(q \cdot \phi_k/\tau)}}
\end{equation}
where $\phi_k$ is the unique representation vector of the $k$-th cluster.
It updates the feature vectors and computes the loss both in the cluster level.

We can see that, our proposed algorithm uses unique feature vectors to represent each cluster category and remains distinct throughout the updating process, which is the most significant difference from the previous contrastive loss approaches.
In the next section, we will discuss in detail how our method consistently updates the cluster representation to maintain the cluster consistency with the help of momentum update.

\subsection {Momentum Update}
\label{sec:ClusterContrast}
In this section, we present how to initialize and update the cluster level memory in the proposed Cluster Contrast method. The training details are presented in Algorithm~\ref{algorithm}.

\paragraph{Memory Initialization.}
Different from the instance level memory dictionary, we store each cluster's representation $\{\phi_1, \dots, \phi_K\}$ in the memory-based feature dictionary. 
We use the mean feature vectors of each cluster to initialize the cluster representation, that is
\begin{equation}\label{eq:ClusterInit_c}
    \phi_k = \frac{1}{|\mathcal{H}_k|}\sum_{u_i \in \mathcal{H}_k}u_i
\end{equation}
Eq.~\ref{eq:ClusterInit_c} is executed when each epoch is initialized. And the clustering algorithm runs in each epoch, so $K$ is changing as the model trains.

\paragraph{Memory Updating.}

During training, following~\cite{hermans2017defense}, $P$ person identities and a fixed number $Z$ of instances for each person identity were sampled from the training set.
Consequently, we obtain a total number of $P \times Z$ query images in the minibatch.
We then momentum update the cluster representation iteratively by the query features in the minibatch by the Eq.~\ref{eq:ClusterUpdate_c} as illustrated in Figure~\ref{fig:loss}:
\begin{equation}\label{eq:ClusterUpdate_c}
    \forall{q}\in Q_k, \phi_k \leftarrow m \phi_k + (1-m) q
\end{equation}
where $Q_k$ is the query features encoded from k-th cluster images and $m$ is the momentum updating factor. $m$ controls the consistency between the cluster feature and most updated query instance feature. As $m$ close to 0, the cluster feature $\phi_k$ is close to the newest query feature. It is worth noting that all cluster representations are stored in the memory dictionary, so we calculate loss by comparing each query instance with all cluster representations in each iteration.

\begin{algorithm}[!t]
    \textbf{Require:} Unlabeled training data $X$\\
    \textbf{Require:} Initialize the backbone encoder $f_\theta$ with ImageNet-pretrained ResNet-50 \\
    \textbf{Require:} Temperature $\tau$ for Eq.~\ref{eq:ClusterNCELoss} \\
    \textbf{Require:} Momentum $m$ for Eq.~\ref{eq:ClusterUpdate_c} \\
    \For{$n$ in $[1, num\_epochs]$}{
        Extract feature vectors $U$ from $X$ by $f_\theta$ \\
        Clustering $U$ into $K$ clusters with DBSCAN \\
        Initialize memory dictionary with Eq.~\ref{eq:ClusterInit_c} \\
        \For{$i$ in $[1, num\_iterations]$}{
            Sample $P \times K$ query images from $X$ \\
            Compute ClusterNCE loss with Eq.~\ref{eq:ClusterNCELoss} \\
            Update cluster feature with Eq.~\ref{eq:ClusterUpdate_c} \\
            Update the encoder $f_\theta$ by optimizer 
        }
    }
    \caption{Unsupervised learning pipeline with Cluster Contrast}
    \label{algorithm}
\end{algorithm}


\section{Experiment}
\begin{table}[!t]             
    \centering
    \caption{Statistics of the datasets used in the experimental section.}
    \label{tab:dataset}
    \resizebox{1.0\linewidth}{!}{
    \begin{tabular}{c|c|c|c|c|c|c}
        \hline
        Dataset & Object & \#train IDs & \#train images & \#test IDs & \#query images & \#total images \\
        \hline
        Market-1501 & Person & 751 & 1 2,936 & 750 & 3,368 & 32,668 \\
        MSMT17 & Person & 1,041 & 32,621 & 3,060 & 11,659 & 126,441\\
        PersonX & Person & 410 & 9,840 & 856 & 5,136 & 45,792 \\
        VeRi-776 & Vehicle & 575 & 37,746 & 200 & 1,678 & 51,003 \\
        \hline
    \end{tabular}}
\end{table}

\begin{table}[!htb]
    \caption{Comparison with state-of-the-art methods on the object re-ID benchmarks. The purely unsupervised methods, unsupervised domain adaptation methods, and camera-aware unsupervised methods are considered for the comparison. The unsupervised domain adaptation method uses additional source domain labeled dataset and the camera-aware method uses the camera ID. 
    }
    \label{tab:benchmarks}
    \subfloat[Experiments on Market-1501 datasets]{
    \resizebox{0.5\linewidth}{!}{
    \begin{tabular}{l|c|cccc}
        \hline
        \multicolumn{1}{c|}{\multirow{2}*{Methods}} &
        \multicolumn{5}{c}{Market-1501}\\
        \cline{2-6}
        &source & mAP & top-1 & top-5 & top-10 \\ 
        \hline
        \multicolumn{6}{l}{\textbf{Purely Unsupervised}} \\
        SSL~\cite{lin2020unsupervised} & None & 37.8 & 71.7 & 83.8 & 87.4 \\
        MMCL~\cite{wang2020unsupervised} & None & 45.5 & 80.3 & 89.4 & 92.3 \\
        HCT~\cite{zeng2020hierarchical} & None & 56.4 & 80.0 & 91.6 & 95.2 \\
        CycAs~\cite{wang2020cycas} & None & 64.8 & 84.8 & - & - \\
        UGA~\cite{wu2019unsupervised} & None & 70.3 & 87.2 & - & - \\
        SPCL~\cite{ge2020self} & None & 73.1 & 88.1 & 95.1 & 97.0 \\
        IICS~\cite{xuan2021intra} & None & 72.1 & 88.8 & 95.3 & 96.9 \\
        OPLG~\cite{zheng2021online} & None & 78.1 & 91.1 & 96.4 & 97.7\\
        RLCC~\cite{zhang2021refining} & None & 77.7 & 90.8 & 96.3 & 97.5 \\
        ICE~\cite{chen2021ice} & None & 79.5 & 92.0 & 97.0 & {\bf 98.1} \\
        PPLR~\cite{cho2022part} & None & 81.5 & 92.8 & 97.1 & 98.1 \\
        {\bf Ours} & None & {\bf 83.0} & {\bf 92.9} & {\bf 	97.2} & 98.0 \\
        \noalign{\vskip 2mm}    
        \hline
        \multicolumn{6}{l}{\textbf{Unsupervised Domain Adaptation}} \\
        MMCL~\cite{wang2020unsupervised} & Duke & 60.4 & 84.4 & 92.8 & 95.0 \\
        AD-Cluster~\cite{zhai2020ad} & Duke & 68.3 & 86.7 & 94.4 & 96.5 \\
        MMT~\cite{ge2020mutual} & MSMT17 & 75.6 & 89.3 & 95.8 & 97.5 \\
        SPCL~\cite{ge2020self} & MSMT17 & 77.5 & 89.7 & 96.1 & 97.6 \\
        TDR~\cite{isobe2021towards} & Duke & 83.4 & 94.2 & - & - \\
        \noalign{\vskip 2mm}    
        \hline
        \multicolumn{6}{l}{\textbf{Camera-aware Unsupervised}}\\
        CAP~\cite{wang2021camera} & None & 79.2 & 91.4 & 96.3 & 97.7 \\
        ICE(aware)~\cite{chen2021ice} & None & 82.3 & 93.8 & 97.6 & 98.4 \\
        PPLR(aware)~\cite{cho2022part} & None & 84.4 & 94.3 & 97.8 & 98.6 \\
        \hline
    \end{tabular}
    }}
    \subfloat[Experiments on MSMT17 datasets]{
    \resizebox{0.48\linewidth}{!}{
    \begin{tabular}{l|c|cccc}
        \hline
        \multicolumn{1}{c|}{\multirow{2}*{Methods}} &
        \multicolumn{5}{c}{MSMT17}\\
        \cline{2-6}
        &source & mAP & top-1 & top-5 & top-10 \\ 
    \hline
    \multicolumn{6}{l}{\textbf{Purely Unsupervised}} \\
    TAUDL~\cite{li2018unsupervised} & None & 12.5 & 28.4 & - & - \\
    MMCL~\cite{xiao2017joint} & None & 11.2 & 35.4 & 44.8 & 49.8 \\
    UTAL~\cite{li2019unsupervised} & None & 13.1 & 31.4 & - & - \\
    CycAs~\cite{wang2020cycas} & None & 26.7 & 50.1 & - & - \\
    UGA~\cite{wu2019unsupervised} & None & 21.7 & 49.5 & - & - \\
    SPCL~\cite{ge2020self} & None & 19.1 & 42.3 & 55.6 & 61.2 \\
    IICS~\cite{xuan2021intra} & None & 18.6 & 45.7 & 57.7 & 62.8 \\
    OPLG~\cite{zheng2021online} & None & 26.9 & 53.7 & 65.3 & 70.2\\
    RLCC~\cite{zhang2021refining} & None & 27.9 & 56.5 & 68.4 & 73.1 \\
    ICE~\cite{chen2021ice} & None & 29.8 & 59.0 & 71.7 & 77.0 \\
    PPLR~\cite{cho2022part} & None & 31.4 & 61.1 & {\bf 73.4} &  {\bf 77.8} \\
    {\bf Ours} & None & {\bf 33.0} & {\bf 62.0	} & 71.8 & 76.7 \\
    \noalign{\vskip 3mm}    
    \hline
    \multicolumn{6}{l}{\textbf{Unsupervised Domain Adaptation}} \\
    MMCL~\cite{wang2020unsupervised} & Duke & 16.2 & 43.6 & 54.3 & 58.9 \\
    ECN~\cite{zhong2019invariance} & Duke & 10.2 & 30.2 & 41.5 & 46.8\\
    MMT~\cite{ge2020mutual} & Market & 24.0 & 50.1 & 63.5 & 69.3 \\
    SPCL~\cite{ge2020self} & Market & 26.8 & 53.7 & 65.0 & 69.8 \\
    TDR~\cite{isobe2021towards} & Duke & 36.3 & 66.6 & - & - \\
    \noalign{\vskip 2mm}    
    \hline
    \multicolumn{6}{l}{\textbf{Camera-aware Unsupervised}}\\
    CAP~\cite{wang2021camera} & None & 36.9 & 67.4 & 78.0 & 81.4 \\
    ICE(aware)~\cite{chen2021ice} & None & 38.9 & 70.2 & 80.5 & 84.4 \\
    PPLR(aware)~\cite{cho2022part} & None & 42.2 & 73.3 & 83.5 & 86.5 \\
    \hline
    \end{tabular}
    }}

    \subfloat[Experiments on PersonX datasets]{
    \resizebox{0.5\linewidth}{!}{
    \begin{tabular}{l|c|cccc}
        \hline \multicolumn{1}{c|}{\multirow{2}*{Methods}} &
        \multicolumn{5}{c}{PersonX}\\
        \cline{2-6}
        &source & mAP & top-1 & top-5 & top-10 \\ 
    \hline
    MMT~\cite{ge2020mutual} & Market & 78.9 & 90.6 & 96.8 & 98.2 \\
    SPCL~\cite{ge2020self} & Market & 78.5 & 91.1 & 97.8 & 99.0 \\
    SPCL~\cite{ge2020self} & None & 72.3 & 88.1 & 96.6 & 98.3 \\
    {\bf Ours} & None & 84.7 & 94.4 & 98.3 & 99.3 \\
    \hline
    \end{tabular}
    }}
    \subfloat[Experiments on VeRi-776 datasets]{
    \resizebox{0.5\linewidth}{!}{
    \begin{tabular}{l|c|cccc}
        \hline
        \multicolumn{1}{c|}{\multirow{2}*{Methods}} &
        \multicolumn{5}{c}{VeRi-776}\\ 
        \cline{2-6}
        &source & mAP & top-1 & top-5 & top-10 \\ 
    \hline
    MMT~\cite{ge2020mutual} & VehicleID & 35.3 & 74.6 & 82.6 & 87.0 \\
    SPCL~\cite{ge2020self} & VehicleID & 38.9 & 80.4 & 86.8 & 89.6 \\
    SPCL~\cite{ge2020self} & None & 36.9 & 79.9 & 86.8 & 89.9 \\
    PPLR~\cite{cho2022part} & None & {\bf 41.6} & 85.6 & {\bf 91.1} & {\bf 93.4} \\
    {\bf Ours} & None & 40.8 & {\bf 86.2} & 90.5 & 92.8 \\
    \hline
    \end{tabular}
    }}
\end{table}

\begin{table}
    \caption{Our method is more robust to batch size changing.}
    \centering
    \resizebox{0.8\linewidth}{!}{
            \begin{tabular}{c|cc|cc|cc|cc}
            \hline
            {\multirow{2}*{Batch size}}  & \multicolumn{2}{c|}{32} & \multicolumn{2}{c}{64} & \multicolumn{2}{c|}{128} & \multicolumn{2}{c}{256} \\
            \cline{2-9}
            & mAP & Rank-1 & mAP & Rank-1 & mAP & Rank-1 & mAP & Rank-1 \\ 
            \hline
            Baseline  & 62.5& 80.3 & 65.5 & 82.7& 71.5 & 86.7 & 73.1 & 87.3\\
            Ours & 79.0 & 90.4 & 80.2 & 91.2 & 81.8 & 92.0 & 83.0 & 92.9  \\ 
            \hline
            \end{tabular}}
    \label{tab:batchsize}
\end{table}

\subsection{Datasets and Implementation}
\paragraph{Datasets.}
We evaluate our proposed method on three large-scale person re-ID benchmarks: Market-1501~\cite{zheng2015scalable}, MSMT17~\cite{wei2018person}, PersonX~\cite{sun2019dissecting}, and one vehicle ReID dataset, VeRi-776~\cite{liu2016large}. Note that the DukeMTMC-reID~\cite{ristani2016performance} has been taken down for ethic issues. The Market-1501 and MSMT17 are widely used real-world person re-identification tasks.
The PersonX is synthesized based on Unity~\cite{riccitiello2015john}, which contains manually designed obstacles such as random occlusion, resolution, and lighting differences.
To show the robustness of our method, we also conduct vehicle re-identification experiments on the widely used real scene VeRi-776 datasets. The details of these datasets are summarized in Table~\ref{tab:dataset}.

\paragraph{Implementation Details.}
We adopt ResNet-50~\cite{he2016deep} as the backbone encoder of the feature extractor and initialize the model with the parameters pre-trained on ImageNet~\cite{deng2009imagenet}. 
After layer-4, we remove all sub-module layers and add global average pooling (GAP) followed by batch normalization layer~\cite{ioffe2015batch} and L2-normalization layer, which will produce 2048-dimensional features. The Gemeralized-Mean (GeM) pooling~\cite{radenovic2018fine} can further improve the performance, which can be seen in appendix.
During testing, we take the features of the global average pooling layer to calculate the consine similarity.
At the beginning of each epoch, we use DBSCAN~\cite{ester1996density} for clustering to generate pseudo labels. 

The input image is resized 256 x 128 for Market-1501, PersonX and MSMT17 datasets, and 224 x 224 for VeRi-776.
For training images, we perform random horizontal flipping, padding with 10 pixels, random cropping, and random erasing~\cite{zhong2020random}. Each mini-batch contains 256 images of 16 pseudo person identities and each person identity containes 16 images. In the case that a person identity has less than 16 images, images are sampled with replacement to compose 16 images.The momentum value $m$ is set to 0.1 and the loss temperature $\tau$ is set to 0.05 for all datasets. We adopt Adam optimizer to train the re-ID model with weight decay 5e-4. The initial learning rate is set to 3.5e-4 with a warm-up scheme in the first 10 epochs, and then reduced to 1/10 of its previous value every 20 epoch in a total of 50 epochs. 


\begin{figure}[t]
    \centering
    \includegraphics[width=\linewidth]{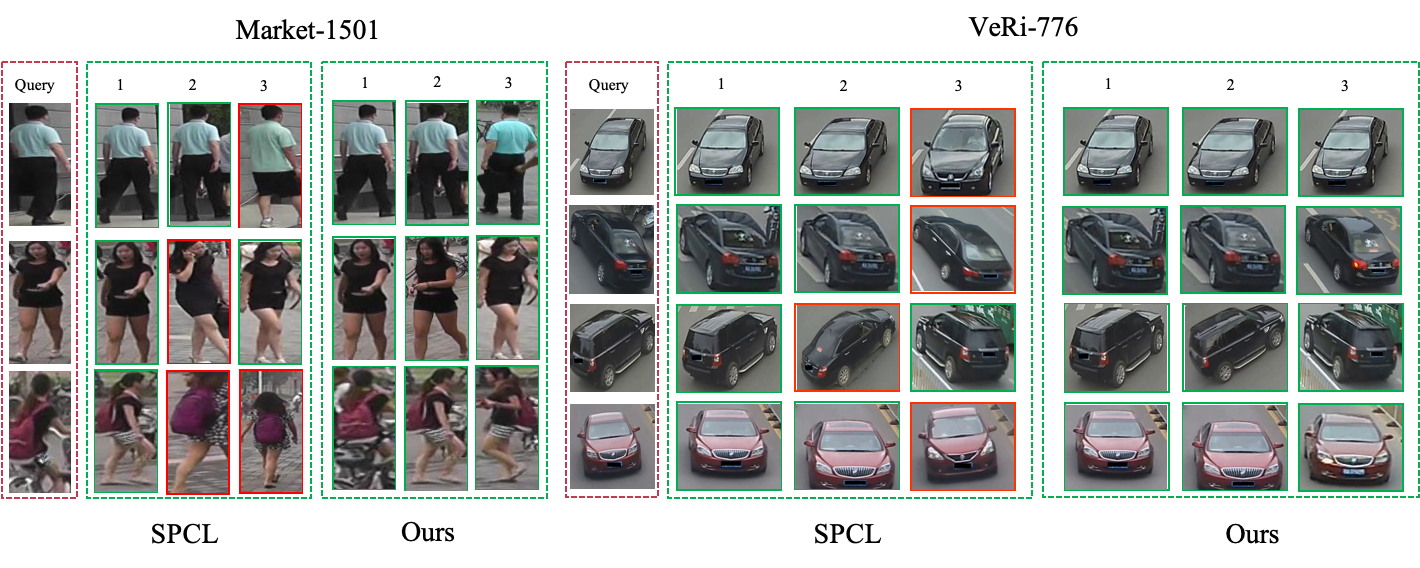}
        \caption{The comparison of top-3 ranking list between SPCL~\cite{ge2020self} and our method on Market1501 and VeRi776 datasets. The correct results are highlighted by green borders and the incorrect results by red borders.}
    \label{fig:query}
\end{figure}

\subsection{Comparison with State-of-the-arts}

We first compare our method to State-of-the-arts unsupervised learning methods which is the main focus of our method. From Table~\ref{tab:benchmarks}, we can see that our method is significantly better than all existing purely unsupervised methods, which proves the effectiveness of our method.
Based on the same pipeline and DBSCAN clustering method, the mAP of our method surpasses the state-of-the-art purely unsupervised learning method by 2.5\%, 2.6\%,  and 12.4\% on person re-ID datasets Market-1501~\cite{zheng2015scalable}, MSMT17~\cite{wei2018person}, and PersonX~\cite{sun2019dissecting} dataset. Our method also performs comparable on vehicle re-ID dataset VeRi-776~\cite{liu2016large}. And our method performs inferior to SOTA UDA and camera-aware unsupervised re-ID methods as they use additional source labeled dataset and camera id information. The Cluster Contrast can be easily generalized on UDA and camera-aware unsupervised re-ID methods. Details can be found in appendix.

\begin{table}[!t]
    \caption{(a) We can see that the cluster level memory remarkably improves the performance and the momentum update strategy can further bring the improvement. (b) The performance is superior when a larger fraction of instances are updated on the baseline method. The statistics of both tables are obtained from the Market1501 dataset.}
    \subfloat[Ablation study of effective components] {
    \label{tab:effect}
    \resizebox{0.5\linewidth}{!}{
        \begin{tabular}{l|cc}
        \hline
        Method & mAP & top-1 \\
        \hline
        Baseline & 73.1 & 87.5 \\
        +Cluster Memory & 80.9 & 91.7 \\
        +Cluster Memory+Momentum & 83.0 & 92.9\\
        \hline
        \end{tabular}
    }
    }
    \subfloat[Impact of cluster size]{
    \label{tab:fraction}
    \resizebox{0.5\linewidth}{!}{
    \begin{tabular}{c|c|c|cc}
        \hline
        Cluster Size & \#Instance & Fraction & mAP & top-1 \\
        \hline
        20 & 4 & 0.2 & 67.6 & 84.4 \\
        8 & 4 & 0.5 & 71.4 & 85.9 \\
        4 & 4 & 1.0 & \textbf{76.5} & \textbf{89.0} \\
        \hline
    \end{tabular}
    }
    }
\end{table}
\begin{figure}[!t]
    \subfloat[Cluster number]{
        \includegraphics[height=38mm]{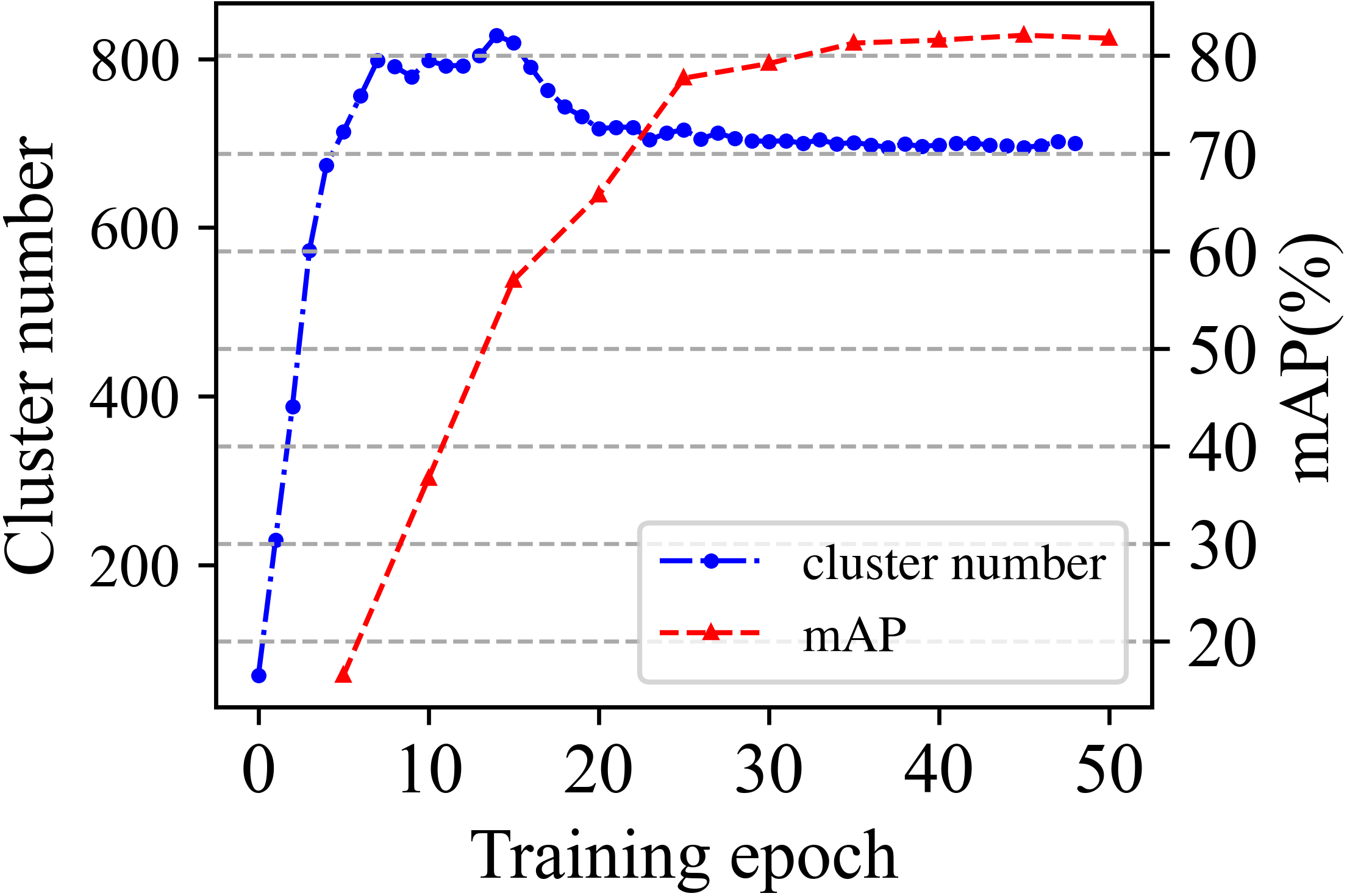}
    }
    \hfill
    \subfloat[~Momentum value]{
        \hspace*{-1cm}\includegraphics[height=38mm]{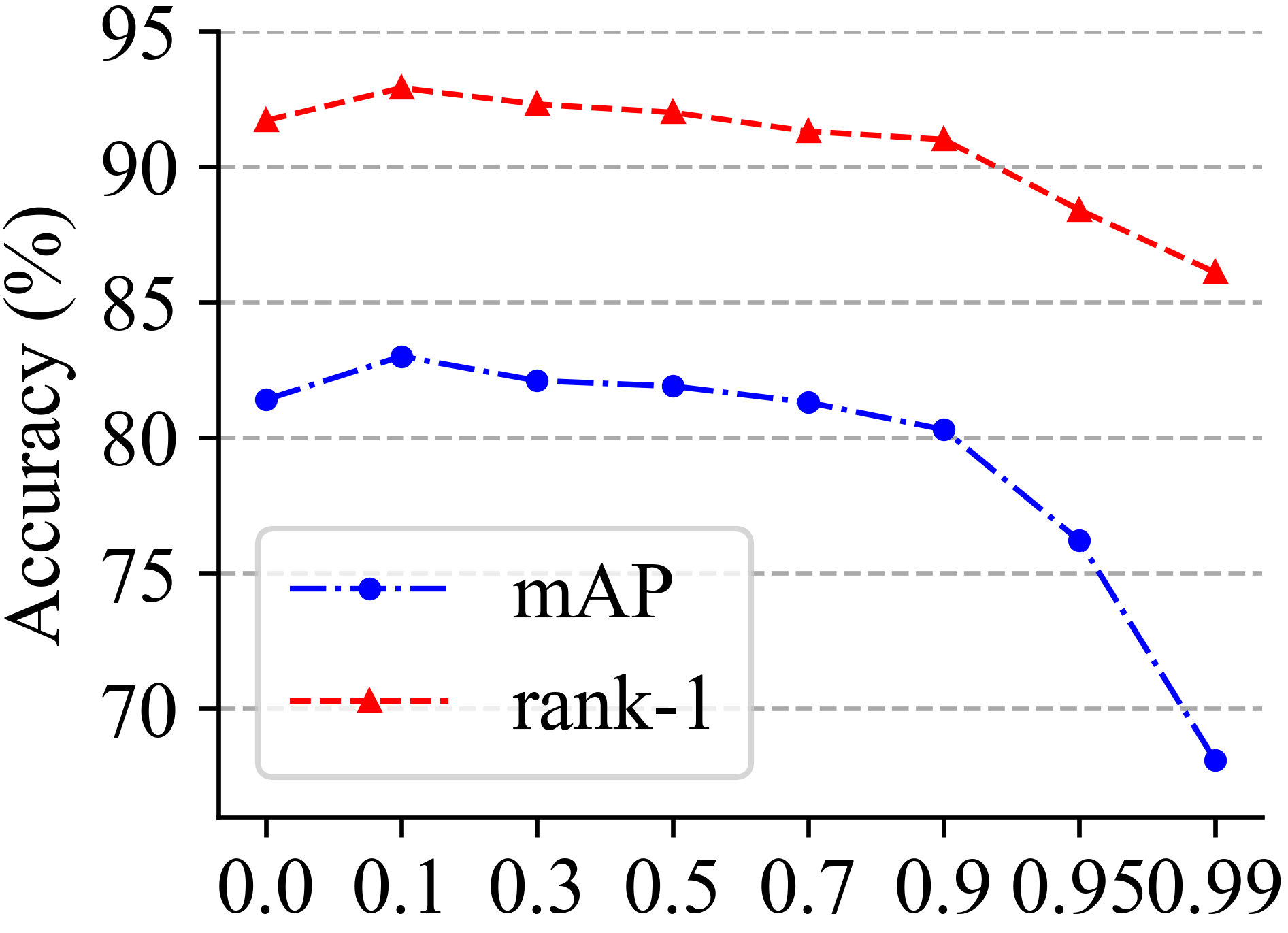}
    }
    \hspace{3mm}
	\caption{(a) We can see that the cluster number is dynamically changing as the model trains. (b) Our method performs comparably well when the momentum value is smaller than 0.9. All the statistics are obtained from the Market-1501.}
	\label{fig:batch}
\end{figure}

\subsection{Ablation Studies}
In this section, we study the effectiveness of various components in Cluster Contrast method. We define the unsupervised learning pipeline with instance-level memory dictionary (Figure~\ref{fig:loss} (b)) as the \textbf{baseline} method. 

\paragraph{Cluster Memory.}
In section~\ref{sec:ClusterContrast}, we argue that compared to instance-level memory, the cluster-level memory could update cluster feature more consistently. As shown in Figure~\ref{fig:loss} (b), the instance-level memory maintains the feature of each instance of the dataset. In every training iteration, each instance feature in the mini-batch will be updated to its own memory dictionary. Since the cluster size is unbalancedly distributed, only a small fraction of the instance features could be updated in a large cluster when all instances in a small cluster are updated. Table~\ref{tab:effect} shows the effectiveness of Cluster Memory without momentum updating, where the instance feature is directly replaced by query feature. 

The simplest solution is to increase the batch size similar to SimCLR~\cite{chen2020simple}. As the batch size increases, more instance features could be updated inside one cluster. However, the batch size reaches its upper limit of 256 due to the GPU memory. To deal with the limitation of the GPU memory, we came up another solution that we restrict the cluster size to a constant number. Therefore, in every iteration a fixed fraction of the instance features could be updated. In this way, the instance feature vectors can be updated consistently with a small batch size. The results in Table~\ref{tab:fraction} demonstrate that the performance of the baseline increases with the rising of the fraction of the updated instance features, until all instance feature vectors inside one cluster could be updated in a single iteration. In sum, we propose the Cluster Contrast, which can update the cluster feature representation in single iteration. As shown in Table~\ref{tab:batchsize}, our method is more robust to batch size changing. And the Cluster Contrast is more memory efficient since the number of cluster features is an order of magnitude
smaller than the number instance features. 

\begin{figure}[!t]
    \centering
    \includegraphics[width=0.7\linewidth]{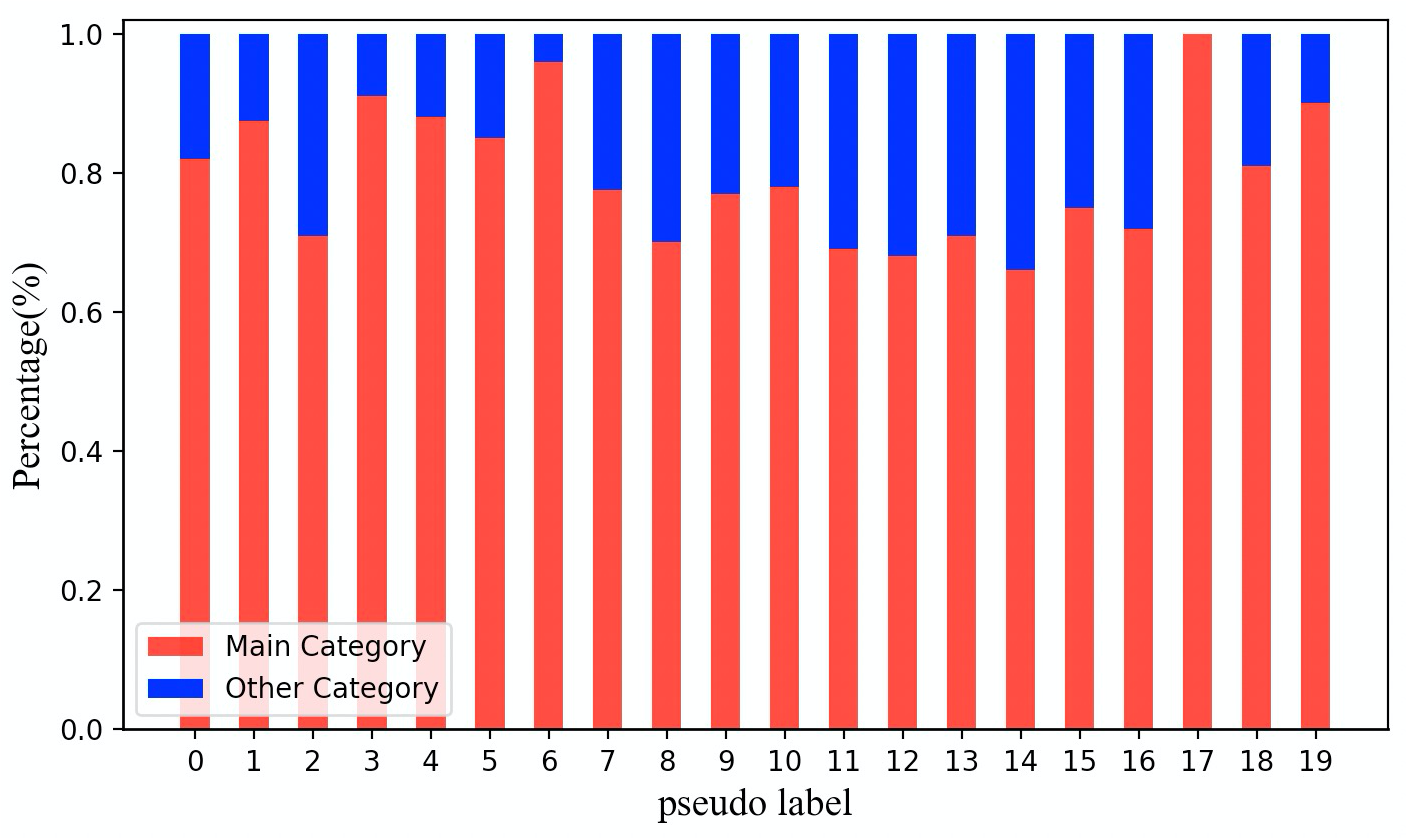}
    \caption{We randomly select 20 categories from the Market1501 clustering results and calculate the percentage of different categories using ground truth labels.}
    \label{fig:-pro}
\end{figure}

\paragraph{Momentum Updating.}
SwAV optimizes the cluster feature end to end by gradient. Our method employs offline clustering so the cluster feature cannot be directly optimized. Instead we follow Moco to use the momentum updating strategy to update cluster representations to maintain the feature consistency. Table~\ref{tab:effect} shows the effectiveness of the momentum updating strategy. As shown in Eq.~\ref{eq:ClusterUpdate_c}, the momentum value $m$ controls the update speed of cluster memory. 
The larger the value of $m$, the slower the cluster memory update. We conducted experiments on the Market-1501 dataset to explore the influence of different $m$ values on our method. As shown in Figure~\ref{fig:batch} (b), it performs reasonably well when $m$ is less than 0.9.  When m is too large (e.g., greater than 0.9), the accuracy drops considerably. These results support us to build better cluster representation.

\begin{figure}[t]
    \subfloat{
        \includegraphics[height=38mm]{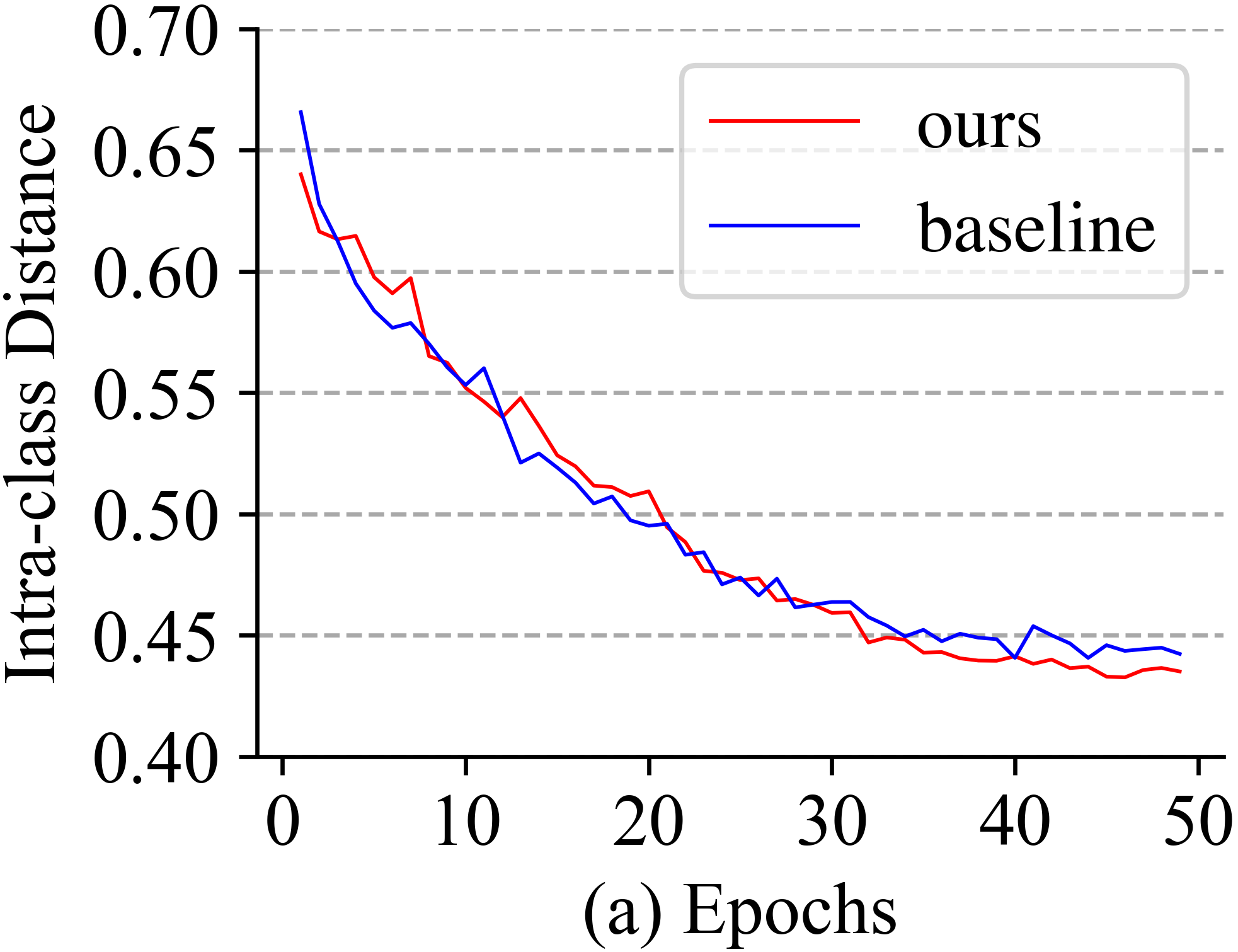}
    }
    \hfill
    \subfloat{
        \includegraphics[height=38mm]{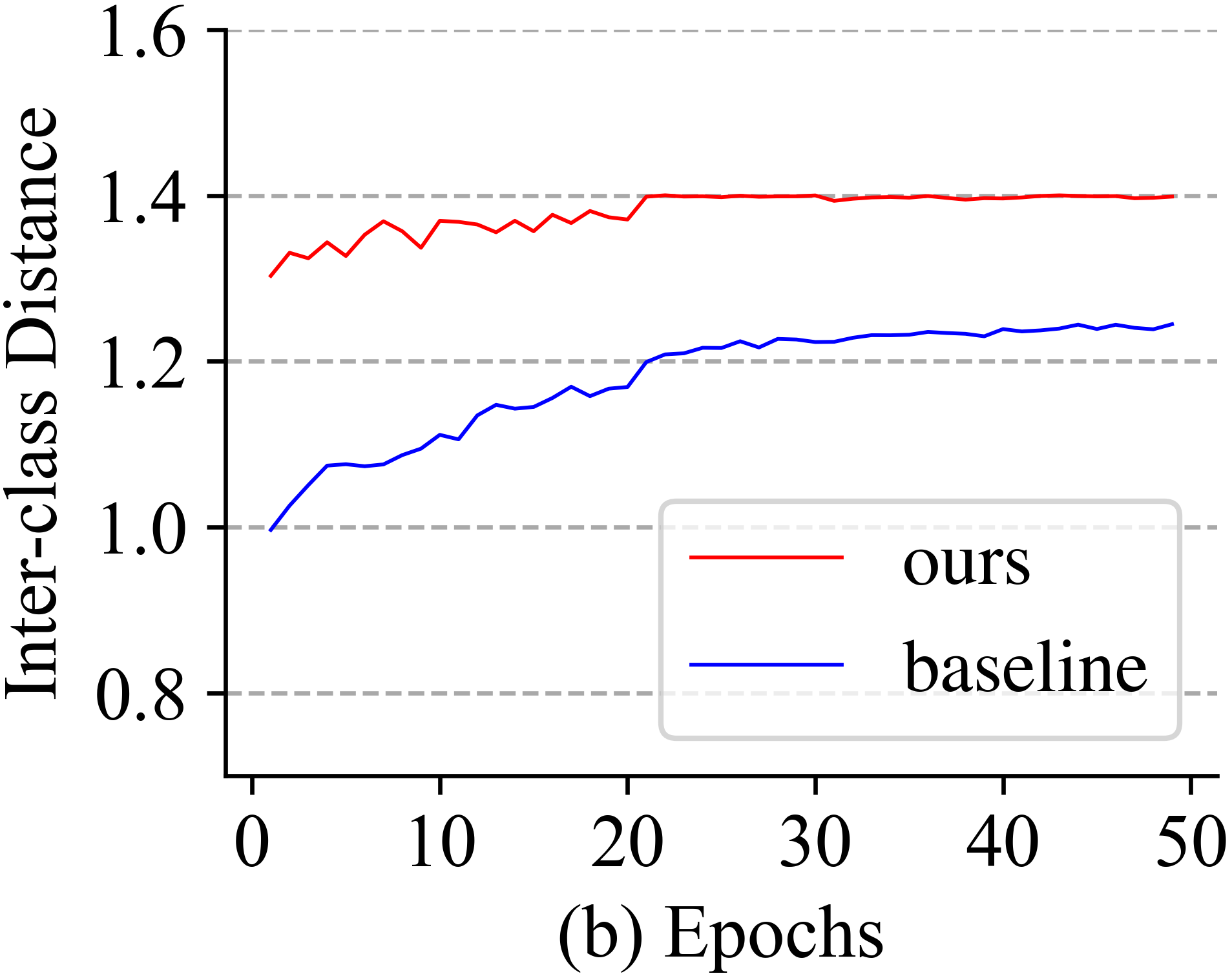}
    }
    \hspace{7mm}
	\caption{Comparison of the intra-class distance and inter-class distance between our method and baseline method on Market1501 datasets.}
	\label{fig:intra_and_inter_distance}
\end{figure}

\paragraph{Cluster Feature Representation.}\label{update}
As shown in Figure~\ref{fig:loss} (b), the instance-level memory averages all instance feature vectors to represent the cluster feature. However, in unsupervised learning re-ID, the pseudo label generation stage would inevitably introduce the outlier instances, which are harmful to compute cluster centroid. In Figure~\ref{fig:-pro}, we count the proportions of different real categories being clustered into the same category on the Market-1501 dataset. It shows there still around 20\% noisy instances when model training in finished. Our method can get better feature representation as shown in Figure~\ref{fig:intra_and_inter_distance}. The feature quality of our method measured by the intra-class distance and the inter-class distance are much better than the baseline method. From this we can speculate that better representation of features between classes is an important factor for our method to achieve better results.


\begin{table}[t]
    \centering
    \caption{Comparison with unsupervised pretrained Resnet50 backbones}
    \label{tab:backbone}
    \begin{tabular}{l|cc|cc|cc|cc}
        \hline
        \multicolumn{1}{c|}{\multirow{2}*{Method}}  &
        \multicolumn{2}{c|}{Market-1501} &
        \multicolumn{2}{c|}{MSMT17} &
        \multicolumn{2}{c|}{PersonX} &
        \multicolumn{2}{c}{VeRi-776} \\
        \cline{2-9}
        & mAP & top-1 & mAP & top-1 & mAP & top-1& mAP & top-1\\ 
        \hline
        supervised & 83.0 & 92.9 & 33.0 & 62.0 & 84.7 & 94.4 & 40.8 & 86.2 \\
        SwAV & \textbf{84.8} & \textbf{93.5} & \textbf{38.2} & \textbf{67.5} & \textbf{86.2} & \textbf{95.1} & \textbf{42.0} & \textbf{87.4} \\
        \hline
    \end{tabular}
\end{table}

\paragraph{Compared with unsupervised pretrained backbones.}
The pseudo-label based unsupervised re-ID methods~\cite{fan2018unsupervised,fu2019self,ge2020self,lin2019bottom,wang2020unsupervised,chen2021ice,wang2021camera,xuan2021intra,zheng2021online,zhang2021refining} use the supervised ImageNet pretrained backbone for clustering initialization. In order to make the pipeline full unsupervised, we also evaluate our method with unsupervised pretrained backbone in Figure~\ref{tab:backbone}. Thanks to the rich feature representation learned by the unsupervised SwAV method, our fully unsupervised re-ID pipeline with unsupervised backbone achieves better results on all four re-ID datasets.




\section{Conclusion}
In this paper, we present the Cluster Contrast for unsupervised re-ID, which stores feature vectors and computes contrast loss in cluster level memory dictionary. It unifies the cluster feature updating progress regardless the cluster size or dataset size. Momentum updating is used to further reinforce the cluster feature consistency. Experiments show demonstrate the effectiveness of our method. 

\clearpage
%
%
%
\bibliographystyle{splncs04}
\bibliography{egbib}
\end{document}


\pagestyle{headings}
\mainmatter

\def\ACCV22SubNumber{250}  

\title{Cluster Contrast for Unsupervised Person Re-Identification} 
\author{Zuozhuo Dai\inst{1} \and
Guangyuan Wang\inst{1} \and
Weihao Yuan\inst{1} \and Siyu Zhu\inst{1} \and Ping Tan\inst{2}}
\authorrunning{Z. Dai et al.}
%
\institute{Alibaba Cloud \and Simon Fraser University}
\maketitle
\section{Appendix}
The Cluster Contrast is simple and effective. It can be used together with existing unsupervised re-ID methods to further improve their performances. In this section, we demostrate the generalization ability of Cluster Contrast by applying it on exsisting methods. For example, the Cluster Contrast can be used with the Generalized Mean Pooling (GeM)~\cite{radenovic2018fine} to further improves the purely unsupervised re-ID performance. In unsupervised domain adaption re-ID, the cluster contrast further boost the SpCL's performance. In particular, With additional camera ID information, the Cluster Contrast with CAP~\cite{wang2021camera} outperforms all SOTA unsupervised re-ID methods.  

\begin{table}
    \centering    
    \caption{The Cluster Contrast further boosts the existing purely unsuprvised re-ID, unsupervised domain adaptation and camera-aware unsupervised re-ID methods}
    \label{tab:cam}
    \resizebox{1.0\linewidth}{!}{
    \begin{tabular}{l|c|cccc|c|cccc}
        \hline
        \multicolumn{1}{c|}{\multirow{2}*{Method}}  &
        \multicolumn{5}{c|}{Market-1501} &
        \multicolumn{5}{c}{MSMT17} \\
        \cline{2-11}
        & source & mAP & top-1 & top5 & top 10 & source & mAP & top-1 &top-5 & top10\\ 
        \hline
        \multicolumn{11}{l}{\textbf{Purely Unsupervised}}\\
        Cluster Contrast & None & 83.0 & 92.9 & 97.2 & 98.0 & None & 33.0 & 62.0 & 71.8 & 76.7 \\
        GeM + Cluster Contrast& None & 84.2 & 93.4 & 97.6 & 98.3 & None & 33.6 & 63.3 & 73.3 & 78.0 \\
        \hline
        \multicolumn{11}{l}{\textbf{Unsupervised Domain Adaptation}}\\
        SPCL~\cite{ge2020self} & MSMT17 & 77.5 & 89.7 & 96.1 & 97.6 & Market & 26.8 & 53.7 & 65.0 & 69.8 \\
        SPCL + Cluster Contrast & MSMT17 & 83.3 & 92.8 & 96.9 & 98.1 & Market & 35.1 & 62.4 & 75.3 & 79.9\\
        \hline
        \multicolumn{11}{l}{\textbf{Camera-aware Unsupervised}}\\
        CAP~\cite{wang2021camera} & None & 79.2 & 91.4 & 96.3 & 97.7& None & 36.9 & 67.4 & 78.0 & 81.4 \\
        ICE~\cite{chen2021ice} & None & 82.3 & 93.8 & 97.6 & 98.4 & None & 38.9 & 70.2 & 80.5 & 84.4\\
        CAP + Cluster Contrast & None & 83.5 & 93.1 & 97.1 & 98.2 & None & 40.1 & 69.8 & 80.2 & 84.0 \\
        \hline
    \end{tabular}
    }
\end{table}  

\subsection{Purely Unsupervised Re-ID}
The Gemeralized-Mean (GeM) pooling~\cite{radenovic2018fine} is a trainable pooling layer which is widely used to improve the image retrieval performance. By replacing the global average pooling with GeM pooling in Resnet-50 backbone, the Cluster Contrast further improves the unsurpervised re-ID performance as shown in Table~\ref{tab:cam}.

\subsection{Unsupervised Domain Adaptation Re-ID}
Our method can be easily generalized to unsupervised domain adaptation re-ID methods. We adopt the hybrid memory from SpCL~\cite{ge2020self} with our method, which combines the  labeled source dataset with the pseudo-labeled target dataset as a unified cluster-level memory. As shown in Table~\ref{tab:cam}, the Cluster Contrast method further boost SpCL's performance.

\subsection{Camera-aware Unsupervised Re-ID}
To solve the large intra-ID variance caused by the change of camera views, the camera aware method~\cite{chen2021ice,wang2021camera} uses the camerea ID information to guide model training. Specifically, they further split the unsupervised clustering results by camera ID and use the camera-aware contrastive loss to pull the positive samples from the same camera closer and push the negtive samples from different cameras away. To show the generalization ability of the Cluster Contrast method, we adopt the cluster-level memory into CAP~\cite{wang2021camera} as an example. As show in Table~\ref{tab:cam}, our method boost the CAP performance significantly, outperforming all unsupervised re-ID methods on MSMT17 dataset.

\clearpage
%
%
\bibliographystyle{splncs04}
\bibliography{egbib}